\documentclass[11pt,a4paper]{article}
\usepackage[hyperref]{acl2021}
\usepackage{times}
\usepackage{latexsym}
\usepackage{graphicx}
\usepackage{multirow}
\usepackage{booktabs}
\usepackage{float}
\usepackage[ruled,vlined]{algorithm2e}
\usepackage{amssymb,amsmath}
\usepackage{graphics}

\usepackage{xcolor}

\usepackage{microtype}

\aclfinalcopy 

\title{Probing Pre-Trained Language Models for Disease Knowledge}

 \author{Israa Alghanmi, Luis Espinosa-Anke, Steven Schockaert \\
         Cardiff University, United Kingdom \\  \texttt{\{alghanmiia,espinosa-ankel,schockaerts1\}@cardiff.ac.uk}}

\date{}

\begin{document}
\maketitle
\begin{abstract}
Pre-trained language models such as ClinicalBERT have achieved impressive results on tasks such as medical Natural Language Inference. At first glance, this may suggest that these models are able to perform medical reasoning tasks, such as mapping symptoms to diseases. However, we find that standard benchmarks such as MedNLI contain relatively few examples that require such forms of reasoning. To better understand the medical reasoning capabilities of existing language models, in this paper we introduce DisKnE, a new benchmark for Disease Knowledge Evaluation. To construct this benchmark, we annotated each positive MedNLI example with the types of medical reasoning that are needed. We then created negative examples by corrupting these positive examples in an adversarial way. Furthermore, we define training-test splits per disease, ensuring that no knowledge about test diseases can be learned from the training data, and we canonicalize the formulation of the hypotheses to avoid the presence of artefacts. This leads to a number of binary classification problems, one for each type of reasoning and each disease. When analysing pre-trained models for the clinical/biomedical domain on the proposed benchmark, we find that their performance drops considerably. 
\end{abstract}

\section{Introduction}
Pre-trained language models (LMs) such as BERT \cite{DBLP:conf/naacl/DevlinCLT19} are currently the de-facto architecture for solving most NLP tasks, and their prevalence in general language understanding tasks is today indisputable \cite{wang2018glue,wang2019superglue}. Beyond generic benchmarks, it has been shown that LMs are also extremely powerful in domain-specific NLP tasks, e.g., in the biomedical domain \cite{lewis2020pretrained}. While there are several reasons why they are preferred over standard neural architectures, one important (and perhaps less obvious) reason is that LMs capture a substantial amount of world knowledge. For instance, several authors have found that LMs are able to answer questions without having access to external resources \cite{DBLP:conf/emnlp/PetroniRRLBWM19,DBLP:conf/emnlp/RobertsRS20}, or that they exhibit commonsense knowledge \cite{DBLP:conf/cogsci/ForbesHC19,DBLP:conf/emnlp/DavisonFR19}. 
To analyze the capabilities of LMs in a more systematic way, there is a growing interest in designing probing tasks, which are now
common across the NLP landscape, e.g., for word and sentence-level semantics \cite{paperno2016lambada,conneau2018you}. In this paper we focus on (generic and specialized) LMs in the biomedical domain, and ask the following question: what kinds of medical knowledge do pre-trained LMs capture? More specifically, we focus on \textit{disease knowledge}, which encompasses for instance the ability to link symptoms to diseases, or treatments to diseases.

Among the several biomedical LMs (i.e.\ LMs that have been pre-trained on biomedical text corpora) that exist today, some of the most prominent are SciBERT \cite{DBLP:conf/emnlp/BeltagyLC19}, BioBERT \cite{lee2020biobert} and ClinicalBERT  \cite{alsentzer2019publicly}. 
Rather than architectural features, these models differ from each other mostly in the pre-training corpora: SciBERT was trained from scratch on scientific papers; BioBERT is an adapted version of BERT \cite{DBLP:conf/naacl/DevlinCLT19}, which was fine-tuned on PubMed articles as well as some full text biomedical articles; and ClinicalBERT was initialized from BioBERT and further fine-tuned on MIMIC-III notes \cite{johnson2016mimic}, which are clinical notes describing patients admitted to critical care units. These  LMs have enabled impressive results on various reading comprehension benchmarks for the medical domain, such as MedNLI \cite{romanov2018lessons} and MEDIQA-NLI \cite{abacha2019overview} for Natural Language Inference (NLI), and PubMedQA \cite{jin2019pubmedqa} for QA. As an example, \citet{wu2019wtmed} achieved an accuracy of 98\% on MEDIQA-NLI, which might suggest that medical NLI is essentially a solved problem. This would be exciting, as medical NLI intuitively requires a wealth of medical knowledge, much of which is not available in structured form.  

However, a closer inspection of MedNLI, the most well-known medical NLI benchmark, reveals three important limitations, namely: (1) only few test instances actually require \textit{medical disease} knowledge, with instances that (only) require terminological and lexical knowledge (e.g.\ understanding acronyms or paraphrases) being more prevalent; 
(2) training and test examples often cover the same diseases, and thus it cannot be determined whether good performance comes from the capabilities of the pre-trained LM itself, or from the fact that the model can exploit similarities between training and test examples; and (3) hypothesis-only baselines perform rather well on MedNLI, which shows that this benchmark has artefacts that can be exploited, similarly to general-purpose NLI benchmarks \cite{poliak2018hypothesis}. 

We therefore propose DisKnE (Disease Knowledge Evaluation), a new benchmark for evaluating biomedical LMs. This dataset explicitly addresses the three limitations listed above and thus constitutes a more reliable testbed for evaluating the disease knowledge captured by biomedical LMs. DisKnE is derived from MedNLI and is organized into two top-level categories, which cover instances requiring medical and terminological knowledge respectively. The medical category is furthermore divided into four sub-categories, depending on the type of medical knowledge that is required.


We empirically analyse the performance of existing biomedical LMs, as well as the standard BERT model, on the proposed benchmark. 
Our results show that all the considered LMs struggle with NLI examples that require medical knowledge. We also find that the relative performance of the pre-trained models differs across medical categories, where the best performance is obtained by ClinicalBERT, BioBERT, SciBERT or BERT depending on the category and experimental setting. Conversely, for examples that are based on terminological knowledge, overall performance is much higher, with relatively little difference between different pre-trained models. 
The contributions of this paper are as follows\footnote{All code for reconstructing the dataset and replicating the experiments is available at: \url{https://github.com/israa-alghanmi/DisKnE}. License and access to MedNLI, MEDIQA-NLI and UMLS will be needed. }:

\begin{itemize}
  \item We introduce a new benchmark to assess the disease-centred knowledge captured by pre-trained LMs, organised into categories that reflect the type of reasoning that is needed, and with training-test splits that avoid leakage of disease knowledge.
  \item We analyze the performance of several clinical/biomedical BERT variants on each of the considered categories. We find that all considered models struggle with examples that require medical disease knowledge. 
  \item We find that without canonicalizing the hypotheses, hypothesis-only baselines achieve the best results in some categories. This shows that the original MedNLI dataset suffers from annotation artefacts, even within the set of entailment examples.
\end{itemize}

\section{Related Work \& Background}
\paragraph{Knowledge Encoded in LMs}
There is a rapidly growing body of work that is focused on analyzing what knowledge is captured by pre-trained LMs. A recurring challenge in such analyses is to separate the knowledge that is already captured by a pre-trained model from the knowledge that it may acquire during a task-specific fine-tuning step. A common solution to address this is to focus on zero-shot performance, i.e.\ to focus on tasks that require no fine-tuning, such as filling in a blank \cite{DBLP:conf/emnlp/DavisonFR19,DBLP:journals/tacl/TalmorEGB20}. 
As an alternative strategy, \citet{DBLP:journals/tacl/TalmorEGB20}  propose to analyse the performance of models that were fine-tuned on a small training set. 
Other work has focused on extracting structured knowledge from pre-trained LMs. Early approaches involved manually designing suitable prompts for extracting particular types of relations \cite{DBLP:conf/emnlp/PetroniRRLBWM19}. Recently, however, several authors have proposed strategies that automatically construct such prompts  \cite{DBLP:conf/aaai/BouraouiCS20,DBLP:journals/tacl/JiangXAN20,shin2020autoprompt}. 
Finally, \citet{bosselut2019comet} proposed to fine-tune LMs on knowledge graph triples, with the aim of then using the model to generate new triples.

\paragraph{LMs for Biomedical Text}
As already mentioned in the introduction, a number of pre-trained LMs have been released for the biomedical domain. Several authors have analyzed the performance of these models, and the impact of including different types of biomedical corpora in particular. For instance, \citet{peng2019transfer} proposed an evaluation framework for biomedical language understanding (BLUE). They obtained the best results with a BERT model that was pre-trained on PubMed abstracts and MIMIC-III clinical notes. Another large-scale evaluation of biomedical LMs has been carried out by \citet{lewis2020pretrained}. To evaluate the biomedical knowledge that is captured in pre-trained LMs, as opposed to acquired during training, \citet{jin2019probing} freeze the transformer layers during training. They find that when biomedical LMs are thus used as fixed feature extractors, BioELMo outperforms BioBERT.
Most closely related to our work, \citet{DBLP:conf/emnlp/HeZZCC20} recently also highlighted the limited ways in which biomedical LMs capture disease knowledge. To address this, they proposed a pre-training objective which relies on a weak supervision signal, derived from the structure of Wikipedia articles about diseases. Other authors have suggested to include structured knowledge, e.g.\ from UMLS, during the pre-training stage of BERT-based models \cite{DBLP:journals/corr/abs-2010-10391,hao2020enhancing}. Another strategy is to inject external knowledge into task-specific models (rather than at the pre-training stage), for instance in the form of definitions \cite{lu2019incorporating} or again UMLS \cite{DBLP:conf/emnlp/SharmaSJTGG19}. 
\citet{kearns2019uw} presented a related approach to our work in which they categorize each sentence pair according to the tense and focus (e.g.\ medication, diseases, procedures, location) of the hypothesis, with the aim of providing a detailed examination of MEDIQA-NLI. Based on this categorization, they compare the performance of Enhanced Sequential Inference Model (ESIM) using ClinicalBERT, Embeddings of Semantic Predications (ESP), and cui2vec. However, their analysis was limited to the MEDIAQ-NLI test set, whereas we include entailment examples from the entire MedNLI and MEDIQA-NLI datasets. Moreover, we focus specifically on the ability of LMs to distinguish between closely related diseases, and we move away from the NLI setting to avoid training-test leakage and artefacts.


\paragraph{Adversarial NLI}
Several Natural Language Inference (NLI) benchmarks have been found to contain artefacts that can be exploited by NLP systems to perform well without actually solving the intended task \cite{poliak2018hypothesis,gururangan2018annotation}. In particular, it has been found that strong results can often be achieved by only looking at the hypothesis of a (premise, hypothesis) pair. In response to this finding, several strategies for creating harder NLI benchmarks have been proposed.
One established approach is to create adversarial stress tests \cite{naik2018stress,glockner2018breaking,aspillaga2020stress}, in which synthetically generated examples are created to specifically test for phenomena that are known to confuse NLI models. This may, for instance, involve the use of WordNet to obtain nearly identical premise and hypothesis sentences, in which one word is replaced by an antonym or co-hyponym. In this paper, we rely on a somewhat similar strategy, using UMLS to replace diseases in hypotheses.
As another strategy to obtain hard NLI datasets, 
\citet{nie-etal-2020-adversarial} used human annotators to iteratively construct examples that are incorrectly labelled by a strong baseline model. While the aforementioned works are concerned with open-domain NLI, some work on creating adversarial datasets for the biomedical domain has also been carried out. In particular, \citet{araujo2020adversarial} studied the robustness of systems for biomedical named entity recognition and semantic text similarity, by introducing misspellings and swapping disease names by synonyms. To the best of our knowledge, no adversarial NLI datasets for the biomedical domain have yet been proposed.

\begin{table*}
\centering
\footnotesize
\begin{tabular}{l@{\hspace{5pt}}cp{170pt}@{\hspace{8pt}}p{125pt}}
\toprule
\textbf{Category} & \textbf{\# inst.} & \textbf{Premise} & \textbf{Hypothesis} \\
\midrule
\textit{Symptoms} $\rightarrow$ \textit{Disease} & 112 & The patient developed neck pain while training with increasing substernal heaviness and left arm pain together with sweating. 
&  The patient has symptoms of acute coronary syndrome 
\\
\midrule
\textit{Treatments} $\rightarrow$ \textit{Disease} & 60 & The patient started on Mucinex and Robitussin. 
&  The patient has sinus disease
\\
\midrule
\textit{Tests} $\rightarrow$ \textit{Disease} & 116 & Cardiac enzymes recorded CK 363, CK-MB 33, TropI 6.78 
& The patient has cardiac ischemia
\\
\cmidrule(lr){3-4}
\multicolumn{1}{l}{} & & A large R hemisphere ICH was revealed when the patent had head CT 
& The patient has an aneurysm 
\\
\midrule
\textit{Procedures} $\rightarrow$ \textit{Disease} & 70 & Bloody fluid was removed by pericardiocentesis
& The patient has hemopericardium.
\\
\midrule
\textit{Terminological} & 259 &  The patient has urinary tract infection
&  The patient has a UTI 
\\
\cmidrule(lr){3-4}
\textit{ } &  &  The patient has high blood pressure 
& Hypertension 
\\
\cmidrule(lr){3-4}
\textit{ } &   & Transfusions in the past could be the cause of the patient having hepatitis C  
& The patient has hepatitis C 
\\
\bottomrule
\end{tabular}
\caption{Considered categories of disease-focused entailment pairs. \label{tabCategories}}
\end{table*}

\section{Dataset Construction}
\label{sec:dataset}
In this section, we describe the process we followed for constructing DisKnE. As we explain in more detail in Section \ref{secSelectingEntailmentPairs}, this process involved filtering the entailment instances from the MedNLI and MEDIQA-NLI datasets, to select those in which the hypothesis expresses that the patient has (or is likely to have) a particular target disease. These instances were then manually categorized based on the type of knowledge that is needed for recognizing the validity of the entailment. Section \ref{secNegativeExamples} discusses our strategy for generating negative examples, which were obtained in an adversarial way, by replacing diseases occurring in entailment examples with similar ones. Details of the resulting training-test splits are provided in Section \ref{secTrainingTest}. In a final step, we canonicalize the hypotheses of all examples, as explained in Section \ref{secCanonicalization}.
Note that the benchmark we propose consists of binary classification problems (i.e.\ predicting entailment or not), rather than the standard ternary NLI setting (i.e.\ predicting entailment, neutral, or contradiction), which is motivated by the fact that natural contradiction examples are hard to find when focusing on disease knowledge.

\subsection{Selecting Entailment Pairs}\label{secSelectingEntailmentPairs}
We started from the set of all entailment pairs (i.e.\ premise-hypothesis pairs labelled with the \textit{entailment} category) from the full MedNLI and MEDIQA-NLI datasets. We used MetaMap to find those pairs whose hypothesis mentions the name of a disease, and to retrieve the UMLS CUI (Concept Unique Identifier) code corresponding to that disease. 
We then manually identified those pairs, among the ones whose hypothesis mentions a disease, in which the hypothesis specifically expresses that the patient has that disease. For instance, in this step, a number of instances were removed in which the hypothesis expresses that the patient does not have the disease. The remaining cases were manually assigned to categories that reflect the type of disease knowledge that is needed to identify that the hypothesis is entailed by the premise. The considered categories are described in Table \ref{tabCategories}, which also shows the number of (positive) examples we obtained and illustrative examples\footnote{For data protection reasons, we only provide synthetic examples, which are different from but similar in spirit to those from the original MedNLI dataset.}. 
The primary distinction we make is between examples that need medical knowledge and those that need terminological knowledge. The former category is divided into four sub-categories, depending on the type of inference that is needed. First, we have the \textit{symptoms-to-disease} category, containing examples where the premise describes the signs or symptoms exhibited by the patient, and the hypothesis mentions the corresponding diagnosis. Second, we have the \textit{treatments-to-disease} category, where the premise instead describe medications (or other treatments followed by the patient). The third category, \textit{tests-to-disease}, involves instances where the premise describes lab tests and diagnostic tools such as X-rays, CT scans and MRI. Finally, the \textit{procedures-to-disease} category has instances where the premise describes surgeries and therapeutic procedures that the patient underwent. In the \textit{terminological} category, the disease is mentioned in both the premise and hypothesis, either as an abbreviation, a synonym or within a rephrased sentence.  


\subsection{Generating Examples}\label{secNegativeExamples}
The process outlined in Section \ref{secSelectingEntailmentPairs} only provides us with positive examples. Unfortunately, MedNLI and MEDIQA-NLI contain only few negative examples (i.e.\ instances of the \emph{neutral} or \emph{contradiction} categories) in which the hypothesis expresses that the patient has some disease. For this reason, rather than selecting negative examples from these datasets, we generate negative examples by corrupting the positive examples.
In particular, to generate negative examples, we replace the disease $X$ from a given positive example by other diseases $Y_1,...,Y_n$ that are similar to $X$, but not ancestors or descendants of $X$ in SNOMED CT \cite{donnelly2006snomed}. To identify similar diseases, we have relied on cui2vec \cite{beam2020clinical}, a pre-trained clinical concept embedding that was learned from a combination of insurance claims, clinical notes and biomedical journal articles. Apart from the requirement that the diseases $Y_1,...,Y_n$ should be similar to $X$, it is also important that they are sufficiently common diseases, as including unusual diseases would make the corresponding negative examples too easy to detect. For this reason, we only consider the diseases that occur in the hypothesis of other positive examples as candidates for the negative examples. Specifically, among these set of candidate diseases, we selected the $n=10$ most similar ones to $X$, which were not descendants or ancestors of $X$ in SNOMED CT (as ancestors and descendants would not necessarily invalidate the entailment). This resulted in a total of 4133 examples requiring medical knowledge and 2639 examples requiring terminological knowledge. 


\subsection{Training-Test Splits}\label{secTrainingTest}
Because our focus is on evaluating the knowledge captured by pre-trained language models, we want to avoid overlap in the set of diseases in the training and test splits. In other words, if the model is able to correctly identify positive examples for a target disease $X$, this should be a reflection of the knowledge about $X$ in the pre-trained model, rather than knowledge that it acquired during training. However, any single split into training and test diseases would leave us with a relatively small dataset. For this reason, we consider each disease $X$ in isolation. Let $\mathcal{E}$ be the set of all positive examples, obtained using the process from Section \ref{secSelectingEntailmentPairs}. Furthermore, we write $\mathcal{E}_X$ for the set of those examples from $\mathcal{E}$ in which the target disease in the hypothesis is $X$. Finally, we write $\textit{neg}(X)$ for the set $\{Y_1,...,Y_n\}$ of associated diseases that was selected to construct negative examples, following the process from Section \ref{secNegativeExamples}.

\begin{figure}[t]
\centering
\includegraphics[scale=0.9]{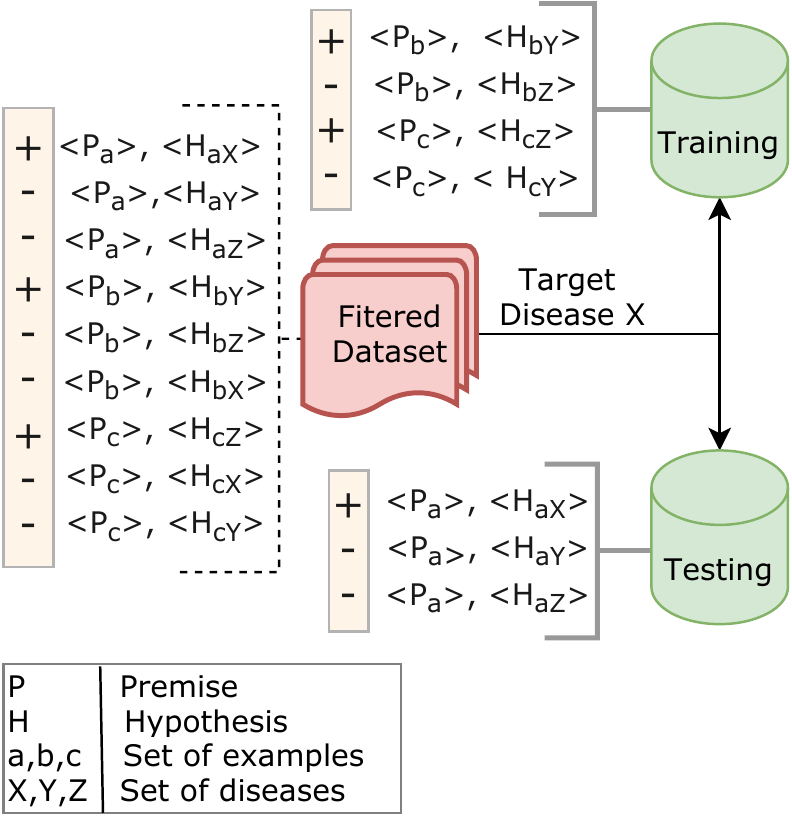}
\caption{Illustration of training-test splitting process.}
\label{fig:Illustration}
\end{figure}

For each target disease $X$, we define a corresponding test set $\textit{Test}_X$ and training set $\textit{Train}_X$ as follows. $\textit{Test}_X$ contains all the positive examples from $\mathcal{E}_X$. Moreover, for each $e\in\mathcal{E}_X$ and each $Y\in \textit{neg}(X)$ we add a negative example $e_{X\rightarrow Y}$ to $\textit{Test}_X$ which is obtained by replacing the occurrence of $X$ by $Y$. If the word before the occurrence of $X$ is \emph{a} or \emph{an}, we modify it depending on whether $Y$ starts with a vowel or consonant. The positive examples in  $\textit{Train}_X$ consist of all examples from $\mathcal{E}$ in which $X$ is not mentioned.
Note that we also remove examples in which these diseases are only mentioned in the premise. Furthermore, we check for occurrences of all the synonyms of these diseases that are listed in UMLS. The process of creating the training and test set for a given target disease $X$ is illustrated in Figure \ref{fig:Illustration}. 

\subsection{Canonicalization}\label{secCanonicalization}
We noticed that the way in which a given hypothesis expresses that ``the patient has disease $X$'' is correlated with the type of the disease. For this reason, as a final step, we canonicalize the hypotheses in the dataset. Specifically, we replace each hypothesis by the name of the corresponding disease $X$. Several hypotheses in the dataset already have this form. By converting the other hypotheses in this format, we eliminate any artefacts that are present in their specific formulation. 

\begin{table}[t]
\footnotesize
\centering
\begin{tabular}{@{}l@{\hspace{4pt}}c@{\hspace{4pt}}c@{\hspace{4pt}}c@{\hspace{4pt}}c@{}}
\toprule
& \rotatebox{90}{\textbf{ClinicalBERT}} & \rotatebox{90}{\textbf{BioBERT}} & \rotatebox{90}{\textbf{SciBERT}} & \rotatebox{90}{\textbf{BERT}} \\
\midrule
\textit{coronary atherosclerosis}	& 0 & 0 & 29 & 10 \\ 
\textit{chf} & 67 & 67 & 67 & 67 \\ 
\textit{acs}	& 04 & 33 & 0 & 05 \\ 
\textit{stroke}	& 80 & 56 & 90 & 90 \\ 
\textit{heart disease}	& 80 & 87 & 93 & 100 \\  
\textit{myocardial infarction}	& 0 & 0 & 19 & 0 \\ 
\textit{heart failure}	& 0 & 0 & 22 & 0 \\
\textit{urinary tract infection\ }	& 100 & 100 & 67 & 100 \\ 
\textit{disorder of lung}	& 89 & 97 & 97 & 100 \\ 
\textit{cirrhosis of liver}& 0 & 11 & 0 & 0 \\ 
\textit{hyperglycemic disorder}& 27 & 13 & 22 & 0 \\ 
\textit{pneumonia}& 89 & 93 & 67 &  100 \\ 
\textit{neurological disease}	& 67 & 67 & 80 & 67 \\ 
\textit{respiratory failure}	& 87 & 70 & 22 & 43\\
\textit{pulmonary edema}	& 74 & 25 & 0 & 50 \\ 
\textit{ami} & 0 & 0 & 0 & 0 \\ 
\textit{deep vein thrombosis} & 47 & 48 & 50 & 48 \\ 
\textit{acute cardiac ischemia} & 0 & 45 & 17  & 72 \\ 
\textit{uri} & 78 & 45 & 67 & 83 \\
\textit{cholangitis} & 22 & 22 & 33 & 22 \\
\textit{atherosclerosis} & 66 & 0 & 67 & 0 \\
\midrule
\textit{Macro-average} & 46\textsubscript{$\pm3.0$} & 42\textsubscript{$\pm7.3$} & 43\textsubscript{$\pm3.1$} & 46\textsubscript{$\pm3.4$} \\ 
\textit{Weighted average} & 49\textsubscript{$\pm3.1$} & 47\textsubscript{$\pm6.0$} & 49\textsubscript{$\pm2.7$ } & 51\textsubscript{$\pm2.7$} \\ 
\bottomrule
\end{tabular}
\caption{Results for the $\textit{Symptoms}\rightarrow\textit{Disease}$ category in terms of F1 (\%) averaged over three runs. Standard deviations (over the three runs) of the macro and weighted average are also reported.}

\label{tabSymptoms}
\end{table}

\begin{table}[t]
\footnotesize
\centering
\begin{tabular}{@{}l@{\hspace{4pt}}c@{\hspace{4pt}}c@{\hspace{4pt}}c@{\hspace{4pt}}c@{}}
\toprule
& \rotatebox{90}{\textbf{ClinicalBERT}} & \rotatebox{90}{\textbf{BioBERT}} & \rotatebox{90}{\textbf{SciBERT}} & \rotatebox{90}{\textbf{BERT}}\\
\midrule
\textit{chf}	& 55 & 55 & 53 & 55 \\
\textit{acs}	& 12 & 19 & 0 & 0 \\
\textit{hypertensive disorder}	& 55 & 67 & 54 & 22\\
\textit{heart disease}	& 45 & 22 & 0 & 89 \\
\textit{urinary tract infection }	& 100 & 100 & 100 & 100 \\
\textit{disorder of lung}	& 82 & 89 & 100 & 93 \\
\textit{hyperglycemic disorder}	& 100 & 69 & 87 & 69\\
\textit{pneumonia}	& 60 & 67 & 78 & 57 \\
\textit{anemia}	& 17 & 17 & 45 & 22\\
\textit{renal insufficiency}	& 69 & 89 & 67 & 72\\
\textit{pulmonary infection}	& 82 & 77 & 89 & 83\\
\textit{copd}	& 45 & 67 & 61 & 39\\
\textit{hyperlipidemia}	& 59 & 61 & 61 & 55 \\
\midrule
\textit{Macro-average} & 60\textsubscript{$\pm6.1$} & 61\textsubscript{$\pm1.4$} & 61 \textsubscript{$\pm3.8$}& 58\textsubscript{$\pm1.6$}\\ 
\textit{Weighted average} & 51 \textsubscript{$\pm5.3$} & 54 \textsubscript{$\pm1.6$}& 51\textsubscript{$\pm1.7$} & 45\textsubscript{$\pm2.4$} \\ 
\bottomrule
\end{tabular}
\caption{Results for the $\textit{Treatments}\rightarrow\textit{Disease}$ category in terms of F1 (\%) averaged over three runs. Standard deviations (over the three runs) of the macro and weighted average are also reported.\label{tabTreatments}}
\end{table}

\section{Experiments}
We experimentally compare a number of pre-trained biomedical LMs on our proposed DisKnE benchmark. In Section \ref{secExperimentalSetup}, we first describe the considered LMs and the experimental setup. The main results are subsequently presented in Section \ref{secResults}. This is followed by a discussion in Section \ref{secDiscussion}.

\subsection{Experimental Setup}\label{secExperimentalSetup}
\paragraph{Pre-trained LMs.} To understand to what extent the pretraining data of an LM affects its performance on our fine-grained evaluation of disease knowledge, we used the following BERT variants:
\begin{description}
\item[BERT] We use the BERT\textsubscript{base}-cased model \cite{DBLP:conf/naacl/DevlinCLT19}.
\item[BioBERT]
\citet{10.1093/bioinformatics/btz682} proposed a model based on BERT\textsubscript{base}-cased, which they further trained on biomedical corpora. We use the version where PubMed and PMC were utilized for this further pre-training. 
\item[ClinicalBERT] \citet{alsentzer2019publicly} introduced four BERT model variants, trained on various clinical corpora. We use the version that was initialized from BioBERT and trained on MIMIC-III notes afterwards.  
\item[SciBERT] \citet{DBLP:conf/emnlp/BeltagyLC19} introduced a BERT model variant that was trained from scratch on approximately 1.14M scientific papers from semantic scholar, 82\% of which were biomedical articles. The full text of the papers was used for training. We use the cased version.
\end{description}

\paragraph{Training Details.}
For fine-tuning, model hyper-parameters were the same across all BERT variants such as the random seeds, batch size and the learning rate. 
In this study,  we fix the the learning rate at 2e-5, batch size of 8 and we set the maximum number of epochs to 8 with the use of early stopping. We used 10\% of the training set as validation split.


\paragraph{Evaluation Protocol.}
We analyze the results per disease and per category in terms of F1 score for the positive class, 
reporting results for all diseases that have at least two positive examples for the considered category. To this end, for each disease $X$, we start from its corresponding training-test split, which was constructed as explained in Section \ref{secTrainingTest}. To show the results for a particular category, we remove from the test set all the examples that do not belong to that category. 

\begin{table}[t]
\footnotesize
\centering
\begin{tabular}{@{}l@{\hspace{4pt}}c@{\hspace{4pt}}c@{\hspace{4pt}}c@{\hspace{4pt}}c@{}}
\toprule
& \rotatebox{90}{\textbf{ClinicalBERT}} & \rotatebox{90}{\textbf{BioBERT}} & \rotatebox{90}{\textbf{SciBERT}} & \rotatebox{90}{\textbf{BERT}} \\
\midrule
\textit{coronary atherosclerosis}	& 0 & 0 & 0 & 0 \\
\textit{chf}	& 52 & 55 & 52 & 55 \\
\textit{acs}	& 0 & 22 & 0 & 0 \\
\textit{stroke}	& 87 & 87 & 95 & 77 \\
\textit{hypertensive disorder}	& 09 & 26 & 45 & 21 \\
\textit{myocardial infarction}	& 28 & 0 & 30 & 14 \\
\textit{heart failure}	& 0 & 55 & 40 & 0 \\
\textit{urinary tract infection} & 87 & 90 & 59 & 90 \\
\textit{hyperglycemic disorder}	& 81 & 10 & 68 & 33 \\
\textit{pneumonia}	& 100 & 100 & 89 & 89 \\
\textit{anemia}	& 0 & 0 & 24 & 0 \\
\textit{aortic valve stenosis}	& 11 & 24 & 0 & 27 \\
\textit{syst.\ inflam.\ resp.\ syndr.\ } & 76 & 64 & 80 & 80 \\
\textit{acute renal failure syndr.\ }	& 0 & 0 & 0 & 22 \\
\textit{chronic renal insufficiency}	& 0 & 0 & 0 & 0 \\
\textit{kidney disease}	& 22 & 0 & 45 & 0 \\
\textit{ischemia}	& 93 & 100 & 93 & 100 \\
\midrule
\textit{Macro-average} & 38 \textsubscript{$\pm2.4$} & 37\textsubscript{$\pm1.6$} & 42\textsubscript{$\pm3.1$} & 36 \textsubscript{$\pm5.0$}\\ 
\textit{Weighted average} & 31\textsubscript{ $\pm2.6$} & 32\textsubscript{$\pm1.2$} & 37\textsubscript{$\pm1.5$} & 31 \textsubscript{$\pm3.7$}\\ 

\bottomrule
\end{tabular}
\caption{Results for the $\textit{Tests}\rightarrow\textit{Disease}$ category in terms of F1 (\%) averaged over three runs. Standard deviations (over the three runs) of the macro and weighted average are also reported.}\label{tabTests}
\end{table}

\begin{table}[t]
\footnotesize
\centering
\begin{tabular}{@{}l@{\hspace{4pt}}c@{\hspace{4pt}}c@{\hspace{4pt}}c@{\hspace{4pt}}c}
\toprule
& \rotatebox{90}{\textbf{ClinicalBERT}} & \rotatebox{90}{\textbf{BioBERT}} & \rotatebox{90}{\textbf{SciBERT}} & \rotatebox{90}{\textbf{BERT}}  \\
\midrule
\textit{coronary atherosclerosis}	& 0 & 0 & 16 & 0 \\
\textit{heart disease}	& 83 & 74 & 84 & 84 \\
\textit{heart failure}	& 33 & 33 & 50 & 0 \\
\textit{cirrhosis of liver}	& 0 & 0 & 0 & 0 \\
\textit{end stage renal disease}	& 37 & 29 & 70 & 79 \\
\textit{respiratory failure}	& 58 & 27 & 57 & 27 \\
\textit{renal insufficiency}	& 100 & 100 & 93 & 100 \\
\textit{cardiac arrest}	& 100 & 100 & 93 &  100\\
\textit{disorder of resp. syst.}	& 76 & 80 & 80 &71  \\
\textit{peripheral vascular dis.}	& 0 & 0 & 78 & 0 \\
\midrule
\textit{Macro-average} & 49 \textsubscript{$\pm3.2$}& 44\textsubscript{$\pm5.9$} & 62\textsubscript{$\pm3.9$} & 46\textsubscript{$\pm5.0$} \\ 
\textit{Weighted average} & 40\textsubscript{$\pm3.3$} & 36 \textsubscript{$\pm7.4$} & 55 \textsubscript{$\pm5.6$} & 44 \textsubscript{$\pm4.6$} \\ 
\bottomrule
\end{tabular}
\caption{Results for the $\textit{Procedures}\rightarrow\textit{Disease}$ category in terms of F1 (\%) averaged over three runs. Standard deviations (over the three runs) of the macro and weighted average are also reported.}\label{tabProcedure}
\end{table}

\begin{table}[t]
\footnotesize
\centering
\begin{tabular}{@{}l@{\hspace{4pt}}c@{\hspace{4pt}}c@{\hspace{4pt}}c@{\hspace{4pt}}c}
\toprule
& \rotatebox{90}{\textbf{ClinicalBERT}} & \rotatebox{90}{\textbf{BioBERT}} & \rotatebox{90}{\textbf{SciBERT}} & \rotatebox{90}{\textbf{BERT}} \\
\midrule
\textit{anemia}& 95 & 100 & 100 & 93  \\
\textit{aortic valve stenosis}& 100 & 100 & 93 & 100 \\
\textit{carotid artery stenosis}& 50 & 50 & 60 & 50  \\
\textit{coronary atherosclerosis}& 79 & 79 & 76 & 79 \\
\textit{type 2 diabetes mellitus}& 67 & 56 & 64 & 61  \\
\textit{gerd}& 0 & 0 & 0 &  0 \\
\textit{cardiac arrest}& 95 & 97 & 92 & 97 \\
\textit{heart disease}& 100 & 100 & 93 & 80 \\
\textit{heart failure}& 100 & 100 & 100 & 100  \\
\textit{chf}& 19 & 37 & 35 &  36 \\
\textit{hyperglycemic disorder}& 57 & 63 & 80 & 57 \\
\textit{hypertensive disorder}& 84 & 87 & 90 &  84 \\
\textit{acute renal failure synd.\ }& 67 & 67 & 58  & 61 \\
\textit{end-stage renal disease}& 77 & 77 & 78 & 70  \\
\textit{disorder of lung}& 89 & 76 & 70 & 52 \\
\textit{copd }& 100 & 100 & 97 &  100 \\
\textit{myocardial infarction}& 24 & 25 & 25 & 21 \\
\textit{pancreatitis}& 33 & 0 & 22 & 33  \\
\textit{pleural effusion}& 80 & 100 & 100 & 80  \\
\textit{pneumonia}& 89 & 93 & 89 & 66  \\
\textit{pulmonary edema}& 87 & 82 & 56 & 76  \\
\textit{stroke}& 81 & 100 & 71 & 100 \\
\textit{urinary tract infection }& 78 & 77 & 78 & 77 \\
\textit{aaa}& 100 & 96 & 100 & 100 \\
\midrule
\textit{Macro-average} & 73 \textsubscript{$\pm2.7$} & 73\textsubscript{$\pm0.4$}  & 72\textsubscript{$\pm2.5$}  & 70\textsubscript{$\pm3.2$}  \\ 
\textit{Weighted average} & 74\textsubscript{$\pm1.8$}  & 76 \textsubscript{$\pm1.4$} & 75 \textsubscript{$\pm1.3$} & 72\textsubscript{$\pm3.0$}  \\ 
\bottomrule
\end{tabular}
\caption{Results for the terminological category in terms of F1 (\%) averaged over three runs. Standard deviations (over the three runs) of the macro and weighted average are also reported.}\label{tabTerminological}
\end{table}

\begin{table}[t]
\footnotesize
\centering
\begin{tabular}{@{}ll@{\hspace{4pt}}c@{\hspace{4pt}}c@{\hspace{4pt}}c@{\hspace{4pt}}c@{\hspace{4pt}}}
\toprule
& &  \multicolumn{2}{c}{\textbf{Standard}} & \multicolumn{2}{c}{\textbf{Hyp.\ only}}\\
\cmidrule(lr){3-4}\cmidrule(lr){5-6}
& &  \textbf{full} & \textbf{can} & \textbf{full} & \textbf{can}\\
\midrule
\multirow{5}{*}{\rotatebox{90}{\textsc{Macro}}}& $\textit{Symptoms}\rightarrow\textit{Dis.}$ & 48 \textsubscript{$\pm0.7$} & 46\textsubscript{$\pm3.0$} & 47\textsubscript{$\pm4.9$}  & 23\textsubscript{$\pm0.5$} \\ 
& $\textit{Treatments}\rightarrow\textit{Dis.}$ & 64\textsubscript{$\pm4.7$} & 60 \textsubscript{$\pm6.1$}& 65\textsubscript{$\pm2.5$} & 29\textsubscript{$\pm2.1$} \\ 
& $\textit{Tests}\rightarrow\textit{Dis.}$ & 41\textsubscript{$\pm1.7$} & 38\textsubscript{$\pm2.4$} & 44\textsubscript{$\pm2.3$} & 18\textsubscript{$\pm2.0$} \\ 
& $\textit{Procedures}\rightarrow\textit{Dis.}$ & 59 \textsubscript{$\pm4.9$}& 49 \textsubscript{$\pm3.2$}& 52\textsubscript{$\pm2.6$} & 19 \textsubscript{$\pm3.0$}\\
& \textit{Terminological} & 71\textsubscript{$\pm2.3$} & 73\textsubscript{$\pm2.7$} & 39\textsubscript{$\pm1.3$} & 25\textsubscript{$\pm0.4$} \\
\midrule
\multirow{5}{*}{\rotatebox{90}{\textsc{Weighted}}}& $\textit{Symptoms}\rightarrow\textit{Dis.}$ & 54 \textsubscript{$\pm2.9$} & 49\textsubscript{$\pm3.1$} &53\textsubscript{$\pm4.7$}  & 23\textsubscript{$\pm1.3$} \\ 
& $\textit{Treatments}\rightarrow\textit{Dis.}$ & 62\textsubscript{$\pm2.8$} & 51\textsubscript{$\pm5.3$} & 60\textsubscript{$\pm7.1$} & 24\textsubscript{$\pm1.0$} \\ 
& $\textit{Tests}\rightarrow\textit{Dis.}$ & 37\textsubscript{$\pm1.4$} & 31\textsubscript{$\pm2.6$} & 42\textsubscript{$\pm0.2$} & 17\textsubscript{$\pm2.8$} \\ 
& $\textit{Procedures}\rightarrow\textit{Dis.}$ & 54\textsubscript{$\pm6.2$} & 40\textsubscript{$\pm3.3$} & 59\textsubscript{$\pm5.1$ }& 14\textsubscript{$\pm2.0$} \\
& \textit{Terminological} & 71\textsubscript{$\pm1.1$}& 74\textsubscript{$\pm1.8$} & 41\textsubscript{$\pm2.7$} & 22\textsubscript{$\pm0.4$} \\
\bottomrule
\end{tabular}
\caption{Comparison between a variant with the full hypothesis and the proposed canonicalized version. Results are for the ClinicalBERT model in terms of F1 (\%) averaged over three runs. Standard deviations (over the three runs) of the macro and weighted average are also reported.}\label{tabCanVsFull}

\end{table}

\begin{table}[t]
\footnotesize
\centering
\begin{tabular}{@{}ll@{\hspace{4pt}}c@{\hspace{4pt}}c@{\hspace{4pt}}c@{\hspace{4pt}}c@{\hspace{4pt}}c@{}}
\toprule
&& 
\rotatebox{90}{\textbf{ClinicalBERT}} & \rotatebox{90}{\textbf{BioBERT}} & 
\rotatebox{90}{\textbf{SciBERT}} & 
\rotatebox{90}{\textbf{BERT}} \\
\midrule
\multirow{5}{*}{\rotatebox{90}{\textsc{Macro}}} & $\textit{Symptoms}\rightarrow\textit{Dis.}$ & 66\textsubscript{$\pm4.0$} & 56\textsubscript{$\pm3.2$} & 57\textsubscript{$\pm5.2$} & 56\textsubscript{$\pm4.1$} \\
& $\textit{Treatments}\rightarrow\textit{Dis.}$ & 69\textsubscript{$\pm4.3$} & 70\textsubscript{$\pm2.0$} & 76\textsubscript{$\pm4.5$} & 55\textsubscript{$\pm4.8$} \\
& $\textit{Tests}\rightarrow\textit{Dis.}$ &53\textsubscript{$\pm0.9$} & 49\textsubscript{$\pm3.3$}& 52\textsubscript{$\pm1.0$} &  47\textsubscript{$\pm0.6$} \\
& $\textit{Procedures}\rightarrow\textit{Dis.}$ & 60 \textsubscript{$\pm1.8$} &56\textsubscript{$\pm0.8$}  & 76\textsubscript{$\pm2.6$} & 60\textsubscript{$\pm4.5$} \\
& \textit{Terminological} & 77\textsubscript{$\pm0.9$} & 77\textsubscript{$\pm0.6$} &74\textsubscript{$\pm0.6$} & 76\textsubscript{$\pm1.0$} \\
\midrule
\multirow{5}{*}{\rotatebox{90}{\textsc{Weighted}}} & $\textit{Symptoms}\rightarrow\textit{Dis.}$ & 66 \textsubscript{$\pm5.2$}& 59\textsubscript{$\pm3.5$} & 59\textsubscript{$\pm4.1$} & 56\textsubscript{$\pm4.6$} \\
& $\textit{Treatments}\rightarrow\textit{Dis.}$ & 64\textsubscript{$\pm6.2$} & 59\textsubscript{$\pm3.6$} & 68\textsubscript{$\pm4.8$}& 46\textsubscript{$\pm3.1$}\\
& $\textit{Tests}\rightarrow\textit{Dis.}$ & 53 \textsubscript{$\pm0.6$} & 51\textsubscript{$\pm2.4$} & 54\textsubscript{$\pm1.6$}& 43\textsubscript{$\pm4.0$} \\
& $\textit{Procedures}\rightarrow\textit{Dis.}$ & 65\textsubscript{$\pm3.0$}& 58\textsubscript{$\pm1.0$}  & 76\textsubscript{$\pm0.4$}& 67\textsubscript{$\pm4.5$}\\
& \textit{Terminological} & 76 \textsubscript{$\pm1.6$} & 77\textsubscript{$\pm1.0$} & 75 \textsubscript{$\pm0.4$} & 72 \textsubscript{$\pm0.7$}\\
\bottomrule
\end{tabular}
\caption{Results for a variant of our benchmark, in which negative examples were selected at random, in terms of F1 (\%) averaged over three runs. Standard deviations (over the three runs) of the macro and weighted average are also reported.}\label{tabRandomResults}

\end{table}

\subsection{Results}\label{secResults}
The main results are shown in Tables \ref{tabSymptoms}--\ref{tabTerminological}. A number of clear observations can be made. First, the results for the terminological category are substantially higher than the results for the other categories, which suggests that the masked language modelling objective, which is used as the main pre-training task in all the considered LMs, may not be ideally suited for learning medical knowledge. Second, recall that the main difference between the considered biomedical LMs comes from the corpora that were used for pre-training them. As the results for the terminological category (Table \ref{tabTerminological}) reveal, the inclusion of domain-specific corpora does not seem to benefit their ability to model biomedical terminology, as 
similar results for this category are obtained with the standard BERT model, which was pre-trained on Wikipedia and a corpus of books and movie scripts. For the $\textit{Symptoms}\rightarrow\textit{Disease}$ category, 
we see that ClinicalBERT outperforms the other biomedical LMs,
although the standard BERT model actually achieves the best performance overall. The results suggest that ClinicalBERT is better at distinguishing between relatively rare diseases, but that the focus on encyclopedic text benefits BERT for more common diseases. Intuitively, we can indeed expect that the encyclopedic style of Wikipedia focuses more on symptoms of diseases than scientific articles, which might focus more on treatments, procedures and diagnostic tests. This is also in accordance with the findings from \citet{DBLP:conf/emnlp/HeZZCC20}, who obtained promising results with a disease-centric LM pre-training task that relies on Wikipedia. On the $\textit{Procedures}\rightarrow\textit{Disease}$ and $\textit{Tests}\rightarrow\textit{Disease}$ categories, we can see that SciBERT achieves the best results, with a particularly wide margin on the $\textit{Procedures}\rightarrow\textit{Disease}$ category. Finally, for the $\textit{Treatments}\rightarrow\textit{Disease}$ category, the relatively poor performance of BERT stands out, which conforms with the aforementioned intuition that scientific articles put more emphasis on procedures, treatments and tests.  BioBERT achieves the best results, although the performance of the other biomedical LMs is quite similar. 

\subsection{Discussion}\label{secDiscussion}
\paragraph{Which LM model?} Several published works have found ClinicalBERT to outperform the other considered biomedical LMs on biomedical NLP tasks \cite{alsentzer2019publicly, kearns2019uw,hao2020enhancing}.
In our results, however, SciBERT achieves the most consistent performance, clearly outperforming 
ClinicalBERT on the $\textit{Procedures}\rightarrow\textit{Disease}$ and $\textit{Test}\rightarrow\textit{Disease}$ categories, while performing similar to ClinicalBERT on the remaining categories.
However, rather than providing a blanket recommendation for SciBERT, our fine-grained analysis highlights the fact that different models have different strengths. The most surprising finding, in this respect, is the performance of the standard BERT model, which achieves the best results on the  $\textit{Symptoms}\rightarrow\textit{Disease}$ category and performs comparably to BioBERT on several other categories (with  $\textit{Treatments}\rightarrow\textit{Disease}$ being a notable exception).

\paragraph{Dataset Artefacts.}
As already reported by \citet{romanov2018lessons}, the original MedNLI dataset has a number of annotation artefacts, which mean that hypothesis-only baselines can perform well. In our dataset, we tried to address this by only using entailment examples, and creating negative examples by corrupting these. However, without canonicalizing the hypotheses, we found that hypothesis-only baselines were still performing rather well. This is shown in Table \ref{tabCanVsFull}, which summarizes the results we obtained for a version of our dataset without canonicalization, i.e.\ where the full hypotheses are provided, and the canonicalized version, where the hypotheses were replaced by the disease name only. The table shows results for the standard ClinicalBERT model, as well as for a hypothesis-only variant, which is only given the hypothesis. 
As can be seen, without canonicalization, the hypothesis only baseline performs similarly to the full model, even outperforming it in a few cases, with the exception of the \emph{Terminological} category where a clear drop in performance for the hypothesis-only baseline can be seen.
In contrast, for the canonicalized version of the dataset, we can see that the hypothesis only baseline, which only gets access to the name of the disease in this case, under-performs consistently and substantially. Note that the hypothesis-only baseline still achieves a non-trivial performance in most cases, noting that an uninformed classifier that always predicts true would achieve an F1 score of 0.167. However, this simply shows that the model has learned to prefer frequent diseases over rare ones.

\paragraph{Adversarial Examples.}
A key design choice has been to select negative examples from the diseases that are most similar to the target disease. To analyse the impact of this choice, we carried out an experiment in which negative examples were instead randomly selected. As before, we only consider diseases that are present in the dataset, and we ensure that negative examples are not ancestors or descendants of the target disease in SNOMED CT. The results are presented in Table \ref{tabRandomResults}. As expected, the results are overall higher than those from the main experiment. More surprisingly, this easier setting benefits some models more than others. The relative performance of ClinicalBERT in particular is now clearly better, with this model achieving the best results for $\textit{Symptoms}\rightarrow\textit{Disease}$. Furthermore, the standard BERT model now clearly underperforms the biomedical LMs, except for $\textit{Procedures}\rightarrow\textit{Disease}$ where it outperforms ClinicalBERT and BioBERT.

\section{Conclusion}
We have proposed DisKnE, a new benchmark for analysing the extent to which biomedical language models capture knowledge about diseases. Positive examples were obtained from MedNLI and MEDIQA-NLI, by manually identifying and categorizing hypotheses that express that the patient has some disease. Negative examples were selected to be similar to the target disease. To prevent shortcut learning, the hypotheses were canonicalized, such that models only get access to the name of the disease that is inferred. Our empirical analysis shows that existing biomedical language models particularly struggle with cases that require medical knowledge. The relative performance on the different categories suggests that different (biomedical) LMs have complementary strengths.

\bibliographystyle{acl_natbib}
\bibliography{acl2021}


\end{document}